%% file: paper.tex
\documentclass[conference]{IEEEtran}
\IEEEoverridecommandlockouts

\usepackage{cite}
\usepackage{amsmath,amssymb,amsfonts}
\usepackage{algorithm}
\usepackage[noend]{algpseudocode}
\usepackage{graphicx}
\usepackage{textcomp}
\usepackage{xcolor}
\def\BibTeX{{\rm B\kern-.05em{\sc i\kern-.025em b}\kern-.08em
    T\kern-.1667em\lower.7ex\hbox{E}\kern-.125emX}}

\usepackage{caption}
\usepackage{subcaption}
\usepackage{booktabs} 
\usepackage{comment}
\newtheorem{definition}{Definition}[section]
\usepackage{lastpage}
\usepackage{tikz}

\newcommand*\circled[1]{\tikz[baseline=(char.base)]{
  \node[shape=circle,draw,fill=black,text=white,font=\bf,inner sep=0.5pt] (char)
  {\scriptsize#1};
}}

\begin{document}

\title{Federated Intrusion Detection for IoT with Heterogeneous Cohort Privacy}

\author{\IEEEauthorblockN{Ajesh Koyatan Chathoth}
\IEEEauthorblockA{
\textit{University of Pittsburgh}\\
Pittsburgh, PA, USA}
\and
\IEEEauthorblockN{Abhyuday Jagannatha}
\IEEEauthorblockA{
\textit{University of Massachusetts Amherst}\\
Amherst, MA, USA }
\and
\IEEEauthorblockN{Stephen Lee}
\IEEEauthorblockA{
\textit{University of Pittsburgh}\\
Pittsburgh, PA, USA}
}

\maketitle

\input{abstract}


\graphicspath{{figures/}}
\section{Introduction}
\label{sec:intro}
\input{introduction}

\section{Preliminaries}
\label{sec:prelim}
\input{prelim.tex}

\section{Heterogeneous Cohort Privacy}
\label{sec:fed_sgd}

\input{baseline.tex}

\subsection{Cohort based Continual Federated Differential Privacy}
\label{sec:cl_dp}
\input{cl_dp.tex}

\section{Evaluation Methodology}
\label{sec:setup}
\input{setup.tex}

\section{Results}
\label{sec:evaluation}
\input{evaluation.tex}

\section{Discussion and Future Work}
\label{sec:future}
\input{future.tex}

\section{Related Work}
\label{sec:related}
\input{relatedwork.tex}

\section{Conclusion}
\label{sec:conclusion}
\input{conclusion.tex}

\bibliographystyle{IEEEtran}

\bibliography{paper.bib}

\end{document}

%% file: abstract.tex
\begin{abstract}
 
Internet of Things (IoT) devices are becoming increasingly popular and are influencing many application domains such as healthcare and transportation. These devices are used for real-world applications such as sensor monitoring, real-time control. In this work, we look at differentially private (DP) neural network (NN) based network intrusion detection systems (NIDS) to detect intrusion attacks on networks of such IoT devices. Existing NN training solutions in this domain either ignore privacy considerations or assume that the privacy requirements are homogeneous across all users. We show that the performance of existing differentially private stochastic methods degrade for clients with non-identical data distributions when clients' privacy requirements are heterogeneous. We define a cohort-based $(\epsilon,\delta)$-DP framework that models the more practical setting of IoT device cohorts with non-identical clients and heterogeneous privacy requirements. 
We propose two novel continual-learning based DP training methods that are designed to improve model performance in the aforementioned setting. To the best of our knowledge, ours is the first system that employs a continual learning-based approach to handle heterogeneity in client privacy requirements. We evaluate our approach on real datasets and show that our techniques outperform the baselines. We also show that our methods are robust to hyperparameter changes. Lastly, we show that one of our proposed methods can easily adapt to post-hoc relaxations of client privacy requirements.

\end{abstract}

%% file: introduction.tex
IoT devices are becoming increasingly prevalent in our daily lives. However, IoT devices have inherent security vulnerabilities making them a prime target for attackers. Prior work has shown how these devices can be easily compromised~\cite{Alasmary8752028, nguyen2019diot}, which has led to the deployment of network intrusion detection systems (NIDS) that detect any malicious network activity. This provides an early-warning system and enable system administrators to detect compromise.

NIDS systems work by monitoring traffic patterns and detecting any malicious activities within the network. It employs an intrusion detection model to identify traffic patterns that deviate from normal behavior. Recently, deep learning-based techniques have been proposed to train the intrusion detection model, which are trained to classify network traffic and identify the type of attack, if any. Since distributed devices may see different types of attack, to effectively capture the heterogeneity of the devices, recent studies have proposed a federated learning approach, where the model learns a common intrusion pattern that captures the behavior of different IoT devices\cite{nguyen2019diot}. The key benefit of federated approach is that it enables aggregation of intrusion patterns from distributed IoT devices such that data remains local to the devices.

Although a decentralized federated approach remains local to the device, it is not privacy-preserving as information may leak from the trained model~\cite{Kairouz2019}. Recently, Differential Privacy (DP) has emerged as a technique to train models to prevent such leak~\cite{abadi2016deep}. DP mechanisms provide statistical guarantees to privacy by perturbing the data using random noise. 
However, prior work mostly assumes  homogeneous privacy requirements across all users. That is, all users have uniform privacy expectations and thus share an almost equal amount of information. 
However, in a realistic scenario, it is quite probable that different users have different privacy budgets. Moreover, providers can incentivize users to share more data. Thus, it is natural to assume that a user will be willing to share more information and be less conservative about their privacy.

    \begin{figure}
        \centering
        \includegraphics[width=3in]{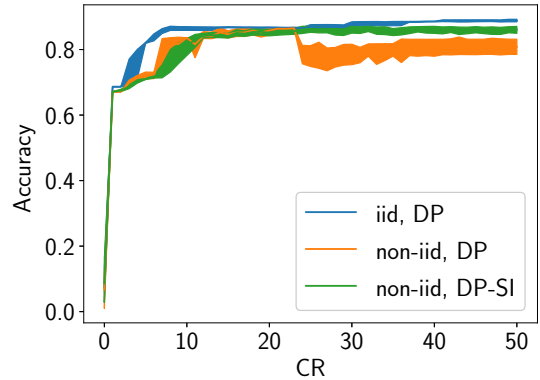}
        \caption{Performance of the model during training for different communication rounds (CR) with heterogeneous cohort privacy. In the non-iid data, DP scenario, the accuracy drops when the privacy budget of users with strict privacy exhausts because the network forgets past experience when new information arrives.}
        \label{fig:IIDvsNon-IID}
    \end{figure}

But, distributed users with heterogeneous privacy budgets where data is not independent and identically distributed
 exacerbate the problem of the training model. In a DP-based deep learning framework, an \emph{accountant} tracks the overall privacy loss at each access to the data to provide a bound on the privacy guarantee~\cite{abadi2016deep}. And, the neural network model training progresses until the privacy budget is exhausted. As shown in Figure~\ref{fig:IIDvsNon-IID}, when the privacy budget is homogeneous (the case assumed in prior work), the performance accuracy remains stable. However, in a heterogeneous privacy budget scenario, the performance drops as it continues to learn. This is because, with new training updates, the model tends to quickly \emph{forget} information learned from past users with stricter privacy budgets. Thus, a key challenge is to ensure that the model does not forget past experiences of users having stricter privacy when new data from users with moderate privacy requirement arrives.

To address the above challenges, in this paper, we design a \emph{continual learning} based approach that ensures that the network remembers past experiences even when the privacy budget of a group of users are spent. In doing so, we make the following key contributions:
\begin{itemize}
    \item We formulate a real world problem of federated learning for intrusion detection on client-cohorts with heterogeneous privacy budgets and non-identical data distribution.  We define the notion of \emph{cohort-based} $(\epsilon,\delta)$-differential privacy for the aforementioned application.
    \item We adapt current federated $(\epsilon,\delta)$-DP training methods for our \emph{cohort-based} $(\epsilon,\delta)$-DP setting and study the challenges introduce by the heterogeneous setting. 
    
    \item We design two novel differentially private continual learning based methods, DP-R and DP-SI, that can effectively train networks with heterogeneous privacy requirements. To the best of our knowledge, this is the first work that uses continual learning to improve federated $(\epsilon,\delta)$-DP training.
    \item  We provide an extensive evaluation that studies the performance, flexibility and hyper-parameter sensitivity of our \textit{cohort-based} federated $(\epsilon,\delta)$-DP SGD methods.  Our evaluation is done on a real world CSE-CIC-IDS2018 dataset \cite{Iman2018}.
    Our results show that continual learning based Federated DP approaches outperform the baseline DP-SGD methods in a heterogeneous privacy setting. The improvement in performance for both our proposed methods is robust to hyperparameter changes. Additionally, we show that DP-SI also provides flexibility in adapting to post-hoc relaxations to client privacy requirements. 
    
\end{itemize}

%% file: prelim.tex
In this section, we provide background on IoT architecture and preliminaries on federated learning and differential privacy. 

\subsection{IoT Architecture and Federated Learning}
 
    \begin{figure}
        \centering
        \includegraphics[width=3.3in]{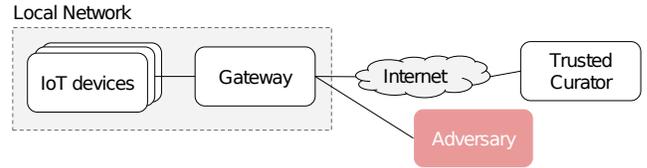}
        \caption{Basic architecture of IoT devices.}
        \label{fig:basic}
    \end{figure}

We consider a three-tier IoT architecture where IoT devices connect to a service provider in the cloud via a gateway. Each user owns a set of IoT devices, and an intrusion detection component within the gateway monitors and detects abnormal communication patterns. The security service provider serves as a curator that trains a central model of anomalous patterns by establishing communication with the distributed gateways. Since a user's data might be unbalanced and non-iid, a key challenge is to learn a model capable of adapting to the device and data heterogeneity. Figure~\ref{fig:basic} depicts the basic architecture assumed in our setup. 

Prior work has looked at a federated learning system for IoT\cite{khan2020federated}. In federated learning, users never share data\cite{mcmahan2017communication}; that is, all data is held locally. This is usually desirable for privacy, security or regulatory reasons. In this scenario, a trusted curator learns a central  model by communicating model parameters with the distributed gateways. As proposed in \cite{mcmahan2017communication}, at each communication round, the trusted curator sends its model parameters to a subset of the gateways. The device gateways perform local optimization and send the model's gradient updates to the service, which then aggregates these updates to improve its central model. The updated model is sent to all gateways, and, the model is finalized after a certain number of communication rounds. 

\subsection{Differential Privacy}

Differential Privacy provides strong guarantees on the information an adversary can infer from a randomized algorithm's output.  Researchers have studied incorporating privacy-preserving mechanisms in federated learning~\cite{geyer2017differentially}. This limits the influence of any single user towards the model's parameters and outputs.

\begin{definition}[$(\epsilon,\delta)$-Differential Privacy]
A randomized mechanism $\mathcal{M}:\mathcal{D} \rightarrow \mathbb{R}$ is $(\epsilon,\delta)$-differential private if for any two neighboring inputs $d_1, d_2 \in \mathcal{D}$ that differ only by a single record, and for all possible outputs ${O} \in \mathbb{R}$: 
    \begin{equation}
 \Pr[\mathcal{M}(d_1) \in O] \le e^{\epsilon} \Pr[ \mathcal{M}(d_2) \in O ] + \delta    
 \label{eqn:dp}
    \end{equation}
\end{definition}
The above definition was introduced by Dwork et. al.~\cite{dwork2014algorithmic}, where the $\epsilon$ parameter is a privacy budget, and controls the privacy. And, a smaller $\epsilon$ value signifies a higher privacy level. 
Whereas $\delta$ is a parameter that controls the probability that the $\epsilon$-differential privacy is broken. The key goal of DP is to limit the contribution of a single data point on the overall output of a function. This contribution is captured using the notion of global sensitivity $S(f)$ and defined as the maximum of the absolute distance.  
\begin{equation}
    S(f) = \max  || f(d_1) - f(d_2) ||_2 
\end{equation} 
where $d_1$ and $d_2$ are neighboring inputs  and $||.||_2$ is the $L_2$ norm. 

Prior work has proposed Gaussian mechanism~\cite{dwork2014algorithmic} that achieves $(\epsilon, \delta)$-DP privacy by calibrating the noise to the sensitivity of the function. The noise is from a Gaussian distribution, and added to the final output of the function as follows $M(d) = f(d) + \mathcal{N}( 0, \sigma^2, S_f^2)$, where $\sigma^2$ is fixed and defines the variance of the Gaussian distribution. Then, an application of GM satisfies $(\epsilon, \delta)$-DP if $\delta \ge \frac{4}{5} \exp(- (\sigma\epsilon)^2/2)$~\cite{dwork2014algorithmic}.
Our work uses this definition of DP. 

\subsection{Differentially Private Learning}
\label{sec:DP_background}
Stochastic gradient descent algorithm (SGD) is a popular optimizer used for training machine learning models. To train a differentially private model, prior work proposed DP-SGD that modifies the SGD algorithm to provide privacy guarantees within the $(\epsilon,\delta)$-DP framework \cite{abadi2016deep}. Intuitively, in DP-SGD, the sensitivity of the gradients are bounded to limit the amount of influence each training samples can have on the model parameters. This is achieved by approximating the gradient averaging step using a Gaussian Mechanism. The basic approach involves sampling a subset of training data, and computing the gradient of the loss with respect to the model parameters. Then, it clips the $l_2$ norm of each gradient, and adds a random noise to the average of the gradients. This average noisy gradient is used to update the model parameters. A privacy \emph{accountant} keeps track of the privacy loss and stops training when the privacy loss reaches a certain threshold. In \cite{abadi2016deep}, the authors proposed the use of higher moments of the privacy loss random variable to track and bound the overall privacy loss. We use the moments accountant framework in \cite{abadi2016deep} to track privacy loss.

\subsection{Definitions}

We now define our notion of \emph{cohort-based} differential privacy, which is an interpretation of the concept of heterogeneous differential privacy~\cite{alaggan2015heterogeneous}. The concept of heterogeneous $(\epsilon,\delta)$-DP  considers privacy at an individual user or item level where the privacy is represented as a vector~\cite{alaggan2015heterogeneous}. In comparison to homogeneous differential privacy, where a single $\epsilon$ value controls the privacy loss, the vector corresponds to each user's privacy.  
In contrast, we define cohort-based privacy that represents the privacy of a collection of users or devices and is defined as follows.

\begin{definition}[Cohort Privacy Mapping]
Let $\mathbb{C}$ denote a set of cohorts and $\mathbb{U}$ denote a set of users. A cohort privacy mapping $\mathcal{G}: \mathbb{U} \rightarrow \mathbb{C}$ maps users to a cohort, where each cohort has the same privacy preference. The mapping $\mathcal{E}: \mathbb{C} \rightarrow \mathbb{R}$ provides the privacy preference of cohort. The notation $\mathcal{E} (\mathcal{G}(u))$ denotes the privacy of the user $u \in \mathbb{U}$.
\end{definition}

Our definition differs from prior approaches in that it provides a more coarser-level definition and use this to set up our contribution in private continual learning. We note that when the size of cohort set $\mathbb{C}$ is equal to the number of users, it decomposes into the scenario where each user picks its own privacy. However, in practice, we expect users to be grouped into cohorts, which provides a more reasonable approach to specifying privacy. We now define cohort differential privacy as follows.

\begin{definition}[Cohort Differential Privacy]
Given the cohort privacy mapping $\mathcal{G}$ and $\mathcal{E}$, 
a randomized mechanism $\mathcal{M}: \mathcal{D} \rightarrow \mathbb{R}$ is said to be cohort differentially private if for all users $u \in \mathbb{U}$, for any two neighboring inputs $d_1, d_2 \in 
\mathcal{D}$ that only differ by a single record $u$, and for all possible outputs $O \in \mathbb{R}$
\begin{equation}
 \Pr[\mathcal{M}(d_1) \in O] \le e^{\mathcal{E} (\mathcal{G}(u) )} \Pr[ \mathcal{M}(d_2) \in O ] + \delta    
 \label{eqn:cohortdp}
 \end{equation}
\end{definition}

We note that the above definition provides privacy guarantee at a cohort level, where the collection of users have same privacy level. It is straightforward to see that when the all the cohorts have the same privacy level, then it is equivalent to the standard $\epsilon$ differential privacy. In our work, we consider each cohort has distinct privacy requirement. 

\subsection{System model and Problem Formulation} 
We consider an environment with multiple IoT devices belonging to users or organizations. We assume users and organizations has capabilities to capture and analyze the packets within their local network. In addition, we also assume that gateways have capabilities to accumulate training data and participate in the federated training. This requires the system to have monitoring features, the ability to capture packets and extract relevant features for modeling. Similar to prior work~\cite{mirsky2018kitsune}, we assume the data is labelled locally, which will be used to train the federated model to detect deviations from anomalous behavior.   

Let $K$ be the number of clients, where clients denote users or organizations. Each client is assigned to a cohort $c \in \mathbb{C}$ from a set of cohorts by a trusted curator. 
Moreover, let $N^i_t$ denote the traffic network packets of the $i^{th}$ client captured at time $t$. 
Given that the privacy requirement at each cohort, our goal is to enable a trusted curator learn an intrusion detection model in a decentralized manner that satisfies the privacy budget within each cohort. Formally, given a sequence of network packets $( N_i^1, N_i^2, ... N_i^t)$, our objective is to classify the network traffic packets and identify the intrusion, if any.

%% file: baseline.tex
Our goal is to train a federated model where users are grouped into cohorts with different privacy requirements.  Additionally, the data distribution for each IoT device and users are not identical. We refer to this setting as the \emph{heterogeneous client scenario}. In what follows, we will describe our approach to design an intrusion detection model within this setup. 

    \begin{figure}
        \centering
        \includegraphics[width=2.4in]{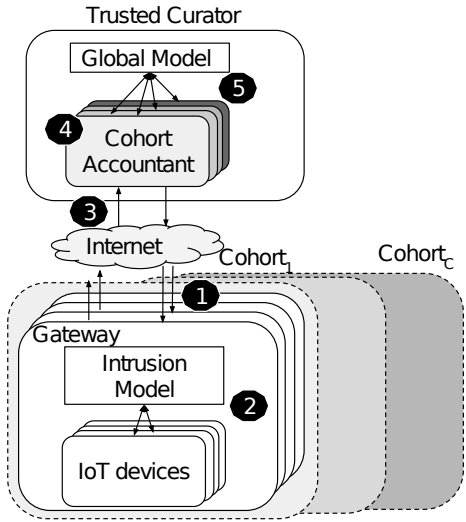}
        \caption{Overview of Cohort based federated training approach.}
        \label{fig:system_overview}
    \end{figure}

\subsection{Cohort based Federated Differential Privacy}
We first modify the federated DP-algorithm to incorporate heterogeneous privacy budgets and adapt the algorithm in \cite{geyer2017differentially} to allow for cohort-specific $\epsilon$ requirements. Our cohort-based ($\epsilon,\delta$)-differential privacy federated algorithm is presented in Algorithm \ref{alg:cohort_alg}.

The key idea is to maintain a cohort-wise privacy accountant instead of a global accountant. And, at each communication round, we track the privacy loss using the moments account~\cite{abadi2016deep}. Figure~\ref{fig:system_overview} illustrates this basic federated model.  IoT devices of each user communicate with a trusted curator to share their local updates. On the other hand, a trusted curator is responsible for assigning users to cohorts and tracking privacy loss. \circled{1} The training starts with the trusted curator sending the global model to a subset of the users in each cohort.  The subset of the user is randomly selected and also helps in amplifying privacy~\cite{balle2018privacy}. \circled{2} After receiving the global model, the model is trained locally by each gateway for a fixed number of steps. \circled{3} The local updates are then transmitted to the cohort privacy accountant within the trusted curator. \circled{4} The cohort privacy accountant aggregates these local updates, adds noise based on the DP mechanism. Moreover, it estimates the privacy loss from client updates. \circled{5}  Finally, the trusted curator aggregates the updates from the cohort accountant and repeats the process until the privacy budget of all the cohorts is exhausted. We note that the accountant evaluates $\delta$ given $\epsilon, \sigma,$ and $m$, where $m$ is a subset of clients participating in a given round. In this case, the $\delta$ value represents the likelihood of a cohort's contribution. The training for a cohort stops when the $\delta$ reaches a certain threshold. For more details, we refer the readers to~\cite{abadi2016deep, geyer2017differentially}.   

Although we incorporate cohort-based accounting in Algorithm \ref{alg:cohort_alg}, we note that it may still provide sub-optimal system utility for the heterogeneous privacy scenario. To investigate why, let us consider a two cohort scenario, where each cohort has clients with a different $\epsilon$ requirement --- one strictly greater than the other. In this scenario, the privacy budget of the cohort with stricter privacy requirements will end sooner than that of the more relaxed cohort.  
If the clients (and cohorts) have nonidentical data distributions, this effectively means that gradient updates from more relaxed cohorts can overwrite the model's information about stricter clients. This phenomenon is referred to as \emph{catastrophic forgetting} \cite{kirkpatrick2017overcoming,lopez2017gradient,li2017learning}. This is also depicted in Figure~\ref{fig:IIDvsNon-IID}. Thus, in a multi-cohort system, \emph{catastrophic forgetting} induced by Algorithm \ref{alg:cohort_alg} can lead to reduced model performance on cohorts with stricter privacy requirements (i.e., lower $\epsilon$ values). We address this problem in the following section.

%% file: cl_dp.tex
\begin{algorithm*}[t]
\caption{DP cohort-wise training. $C$ is the number of cohorts. $K^{(c)}$ is the number of participating clients in each cohort; $B$ is the local mini-batch size, $E$ the number of local epochs, $\eta$ is the learning rate, $\{\sigma\}_{t=0}^{T}$ is the set of variances for the GM. $\{m_t\}_{t=0}^{T}$ determines the number of participating clients at each communication round. $\epsilon$ defines the DP we aim for.  $Q$ is the threshold for $\delta$, the probability that $\epsilon$-DP is broken. $T$ is the number of communication rounds after which $\delta$ surpasses $Q$. $\mathcal{B}$ is a set holding client's data sliced into batches of size B.}
\label{alg:cohort_alg}
\begin{algorithmic}[1]
\Procedure{DP-Server}{}
\For {each cohort $c = 1,2,...C$}
\State Initialize: $w_0, \text{Accountant}(\epsilon_{c}, K_{c})$ \Comment{initialize weights and the privacy accountant} 
\EndFor
\For {each round $t = 1,2,...$}

\For {each cohort $c = 1,2,...C$ in parallel}
\State $\triangle w^{(c)}_{t+1} \gets \text{DP-Cohort($c,w_t$)}$  \Comment{cohort c's update}
\EndFor

\State $w_{t+1} \gets w_t + \frac{1}{C}(\sum_{c=1}^C \triangle w^{(c)}_{t+1}))$ \Comment{Update central model}
\EndFor
\EndProcedure

\Function{DP-Cohort}{$c, w_t$}
\State $\delta \gets \text{Accountant}(m_t^{(c)}, \sigma_t)$ \Comment{Accountant returns priv. loss for current round}
\If {$\delta > Q$} 
return $w_t$ \Comment{If privacy budget is spent, return current model}
\EndIf
\State $Z^{(c)}_t \gets \text{random set of $m_t^{(c)}$ clients} $ \Comment{randomly allocate a set of $m_t^{(c)}$ clients out of $K_{c}$ clients in the cohort c}
\For {each client $k \in Z^{(c)}_t$ in parallel}
\State $\triangle w_{t+1}^k, \zeta^k \gets \text{DP-Client($k,w_t$)}$  \Comment{client k's update and the update's norm}
\EndFor

\State $w^{(c)}_{t+1} \gets w_t + \frac{1}{m^{(c)}}(\sum_{k=1}^{K^{(c)}} \triangle w_{t+1}^k / \text{max}(1,\frac{\zeta^k}{S}) + \mathcal{N}(0,S^2 \cdot \sigma ^2))$ \Comment{Update central model}

\State return $\triangle w^{(c)}_{t+1}$  \Comment{return update and norm of update}
\EndFunction

\Function{DP-Client}{$k, w_t$}

\State $w \gets w_t$
\For {each local Epoch $i = 1,2,...E$}
\For {batch $b \in \mathcal{B}$}
\State $w \gets w - \eta \nabla L(w;b) $ \Comment{mini batch gradient descent}

\EndFor
\EndFor
\State $\triangle w_{t+1} = w - w_t$ \Comment{clients local model update}
\State $\zeta = \Vert \triangle w_{t+1} \Vert _2$ \Comment{norm of update}

\State return $\triangle w_{t+1}, \zeta $  \Comment{return update and norm of update}
\EndFunction

\end{algorithmic}
\end{algorithm*}

\begin{algorithm*}[t]
\caption{DP-R: Rehearsal based continual learning algorithm. $C$ is the number of cohorts. $K^{(c)}$ is the number of participating clients in each cohort, $\rho$ is the rehearsal ratio. $T^{(c)}$ is the number of communication rounds for a cohort $c$ and $S^{(c)}$ indicates the number of communication rounds between each rehearsal rounds for cohort $c$. DP-client and DP-Cohort functions are identical to Algorithm \ref{alg:cohort_alg}}
\label{alg:rehearsal_alg}
\begin{algorithmic}[1]
\Procedure{DP-Rehearsal SERVER}{}
\For {each cohort $c = 1,2,...C$}
\State Initialize: $w_0, \text{Accountant}(\epsilon_{c}, K_{c})$ \Comment{initialize weights and the privacy accountant} 
\State $skip^{(c)} \gets S^{(c)}$ \Comment{initialize the skip length for the communication rounds}
\EndFor
\For {each round $t = 1,2,...$}
\For {each cohort $c = 1,2,...C$ in parallel}
\If {$\text{Rehearsal($t, \rho^{(c)}, T^{(c)}, skip^{(c)}, S^{(c)}$) = 1}$}
\State $\triangle w^{(c)}_{t+1} \gets \text{DP-Cohort($c,w_t$)}$  \Comment{cohort c's update}
\EndIf
\EndFor

\State $w_{t+1} \gets w_t + \frac{1}{C}(\sum_{c=1}^C \triangle w^{(c)}_{t+1}))$ \Comment{Update central model}
\EndFor
\EndProcedure

\Function{Rehearsal}{$t, \rho^{(c)}, T^{(c)}, skip^{(c)}, S^{(c)}$}
\If {$\lceil(1 - \rho^{(c)}) T^{(c)}\rceil > t \textrm{ AND }  skip^{(c)} == 0$} \Comment{if rehearsal stage is reached for the cohort $c$ and this communication round $t$ is a rehearsal round}
\State $skip^{(c)} = S^{(c)}$ \Comment{reset the skip length in rehearsal stage}
\State return $1$ \Comment{return this round $t$ is rehearsal round}

\EndIf
\If {$\lceil(1 - \rho^{(c)}) T^{(c)}\rceil > t$}
\State $skip^{(c)} = skip^{(c)} - 1$ \Comment{decrease the skip length to rehearsal round}
\EndIf
\State return $0$ \Comment{return this round $t$ is not a rehearsal round}

\EndFunction
\end{algorithmic}
\end{algorithm*}

\begin{algorithm*}[t]
\caption{DP-SI: Synaptic Intelligence based continual learning. In addition to variables in algorithm \ref{alg:cohort_alg}, $C$ is the number of cohorts, $\gamma$ is the weight parameter associated with previous task SI loss versus current task loss. $L_{SI}$ is the SI loss from the cohort whose privacy budget exhausted in the most recent round of communication. $\Omega$ is the set of per-parameter regularization strength at the end of each cohort's tasks. DP-client and DP-Cohort functions are identical to Algorithm \ref{alg:cohort_alg}}
\label{alg:SI_alg}
\begin{algorithmic}[1]
\Procedure{DP-SI-Server}{}
\For {each cohort $c = 1,2,...C$}
\State Initialize: $w_0, \text{Accountant}(\epsilon^{c}, K^{c}), L_{SI}=0, j = 0$ \Comment{initialize weights, the privacy accountant and SI loss function and j is the number of cohort's privacy budget exhausted in communication round $t$} 
\EndFor
\For {each round $t = 1,2,...$}

\For {each cohort $c = j+1,j+2,...C$ in parallel in the increasing order of $\epsilon$}
\State $\triangle w^{(c)}_{t+1} \gets \text{DP-Cohort($c,w_t$)}$  \Comment{cohort c's update}

\State $\Omega^{(c)} \gets \text{Update-SI($\triangle w^{(c)}_{t+1},w_t$)} $ \Comment {Update per parameter regularization strength for cohort c}
\If {\textrm{ cohort c's privacy budget is exhausted } }

\State $j \gets j + 1$ \Comment{update the number of cohorts whose privacy budget is exhausted}
\EndIf
\EndFor
\State $L_{SI} \gets \text{Compute-SI-Loss($\mathbf{\Omega},j$)} $ \Comment{Compute SI loss of cohorts whose privacy budget is exhausted before  $t^{th}$ round }


\State $w_{t+1} \gets w_t + \frac{1}{C-j}(\sum_{c=j+1}^C \triangle w^{(c)}_{t+1})) + \gamma \nabla L_{SI}$ \Comment{Update central model}
\EndFor
\EndProcedure
\end{algorithmic}
\end{algorithm*}

To mitigate the effect of forgetting previously learned experiences, we adapt continual learning methods to ensure that model performance does not drop for cohorts with stricter privacy budgets. Continual learning (CL) methods address the problem of improving model performance on a set of tasks that are learned through a sequential training curriculum \cite{mccloskey1989catastrophic,kirkpatrick2017overcoming}. In the case of our two-cohort example, the two tasks are Task $A$: combined training of cohort-1\&2 for the first 23 CRs, and Task $B$: cohort-2 training when cohort-1's CR allowance is exhausted. Throughout this section, we will use this 2 cohort example to explain our proposed methods. However, our proposed algorithms are applicable to any $N$-cohort system. In case of an $N$-cohort system, the continual learning method can be easily extended to $N$ tasks, where each task's training ends at the exhaustion of a specific cohort's CR allowance.

A straightforward application of CL methods is not possible due to the DP requirements in our work. CL methods have not been analyzed with respect to their privacy budgets in any existing work. Therefore, we first study the feasibility of using continual learning methods in our context. We consider the following widely used CL methods: 

\begin{itemize}
    \item \textbf{Elastic Weight Consolidation (EWC)\cite{kirkpatrick2017overcoming}:} EWC uses the diagonal empirical Fisher information matrix of the model parameters. Estimation of empirical fisher diagonals require gradient queries from cohort-1.
    \item \textbf{Gradient Episodic Memory (GEM) \cite{lopez2017gradient,chaudhry2018efficient}: } GEM queries previous task gradients to correct later task gradients. So, GEM also requires new gradient queries from cohort-1.
    \item \textbf{Synaptic intelligence (SI)\cite{zenke2017continual}:} SI requires accumulation of previous task gradients during that task's training procedure. No new gradient queries are required during later task training. So cohort-1 gradients during task $A$ can be reused for SI loss in task $B$.
    \item \textbf{Rehearsal \cite{li2017learning}:} Rehearsal requires cohort-1 gradient for rehearsal loss during subsequent task $B$ training. 
\end{itemize}

We choose Rehearsal and SI as our proposed CL methods for our experiments. Our choice is based on two factors: (i) the additional privacy costs required by each method, and (ii) the potential for model's performance improvements due to \emph{positive forward and backward transfer} \cite{lopez2017gradient}. SI does not require any additional privacy budget because it can reuse gradients required by Federated SGD. The high privacy budget requirements of Rehearsal can be reduced by only periodically accessing the previous task gradients during later task training. 

In contrast, the other two CL methods EWC and GEM require high privacy budgets that cannot be reduced without severely affecting the performance of the method. Reducing the number of gradient queries for EWC would severely affect the reliability of the Fisher information matrix diagonals. Separately, reducing the frequency corrections of GEM would allow gradient steps that have negative components towards previous task gradients. Our continual learning-based Federated DP-SGD algorithms, namely DP-SI and DP-R, are defined in the following sub-sections.

\subsubsection{DP-Synaptic Intelligence (DP-SI)}
Synaptic intelligence (SI) \cite{zenke2017continual} aims to reduce \emph{catastrophic forgetting} of previous tasks by minimizing subsequent changes to model parameters that were \emph{influential} for the previous tasks. This is achieved by adding a quadratic SI loss to the objective function of later tasks,
\begin{equation}
    L^v_{SI}= \sum_{u<v} \sum_{l}\Omega_l^u(\Tilde{\theta}^u_l - \theta_l)^2.
\end{equation}

The equation defines SI loss for task $v$. Here the per-parameter variable $\Omega_l^u$ is a measure of the importance of the parameter $\theta_k$ for task $u$. This importance is determined by factors like parameter change during training and parameter's contribution towards loss reduction. The variable $\Omega_l^u$ is estimated for task $u$ by 

\begin{equation}
    \Omega_l^u=\frac{w_l^{u}}{(\Delta_l^u)^2 + \mathcal{E}}
\end{equation}

Here, $\Delta_l^u$ is the change in $\theta_l$ for task $u$ and $w_l^{u}$ is the running sum of the product of the loss gradient with respect to the parameter and the parameter update value for that task. The damping parameter $\mathcal{E}$ is used to ensure numerical stability in cases where $\Delta_l^u \rightarrow 0$. The detailed sketch of our DP-SI algorithm is provided in Algorithm \ref{alg:SI_alg}. For our client-cohorts experimental setup, without loss of generality we can assume that cohorts are named such that epsilon for cohort-$u$ is less than that of cohort-$(u+1)$. The loss for SI (in a CR after cohort v is exhausted but before $v+1$ is exhausted) is defined as:

\begin{equation}
    L^{v+1} = \sum_{u>v} L^{u}_{cohort} + \gamma \sum_{u\leq v} L^{u}_{SI}. 
\end{equation}

Here, $L^{u}_{cohort}$ is the empirical risk minimization loss on cohort-$u$ data. The hyperparameter $\gamma$ is used to weight the contribution of SI loss. The per-parameter importance weights $\Omega^u_l$ are computed based on gradient information till the communication round allowance for cohort-$u$ is exhausted. The estimation of $\Omega^u_l$ is done on the server-side, using the perturbed client-cohort gradients. The injected Gaussian noise in client gradients, as defined in Algorithm \ref{alg:SI_alg}, provides us with a biased estimate of $w_l$, and as a result, $\Omega_l$ is also biased. However, in practice, our experiments show that these per-parameter importance weights are still useful in reducing the effects of \emph{catastrophic forgetting}. 

The estimation of SI's per-parameter importance only reuses the gradient information that is provided for federated DP-SGD. The immunity to post-processing property of $(\epsilon,\delta)$-DP \cite{Dwork2006a} ensures that reusing the client gradients does not change the privacy leakage. 

\subsubsection{DP-Rehearsal (DP-R)}
We also adopt a widely used continual learning procedure called Rehearsal \cite{li2017learning} for our heterogeneous privacy setting. Unlike SI, Rehearsal cannot reuse client gradients from previous CRs to reduce \emph{catastrophic forgetting}.  It requires gradients from $\{\textrm{cohort-}u\}_{u\leq v}$ data periodically during task $v+1$ training. This periodic rehearsal of previous cohort data also spends the privacy budget. Therefore, a certain fraction of each cohort's privacy budget has to be kept aside in order to allow for periodic rehearsal updates throughout the model training process.

The detailed algorithm is provided in Algorithm \ref{alg:rehearsal_alg}. For simplicity of explanation, let us consider the two cohort example from previous sections. The first phase of the training (task $A$) when cohort-1 privacy budget is used with cohort-2 to train the underlying model, lasts for $\lceil 1-\rho \times T_1\rceil$ CRs. Here, $\rho$ is a hyperparameter that controls the fraction of each cohort-1's privacy budget that is kept aside for rehearsal. This remaining CR allowance is used to periodically query cohort-1 gradient and update the model throughout the rest of task $B$ training. This ensures that the model performance on cohort-1 data does not \emph{catastrophically reduce} during task $B$. 

The periodicity of gradient queries for any cohort is based on $\rho$ and the maximum CRs allowance. To estimate the periodicity for  DP-R, the max CR count must be fixed and known beforehand. This is a major weakness in DP-R as compared to DP-SI, which does not require any knowledge about max CRs.

%% file: setup.tex
We now describe the dataset, our experimental setup, and metrics used to evaluate our approach.  

\subsection{Datasets}
\begin{table}[]
    \centering
    \begin{tabular}{l|c}
        \toprule
         Attack scenario & Number of data samples \\ \midrule
         Benign&  671244 (67\%) \\ 
         Bot& 66308 (6.7 \%) \\
         DOS attacks-GoldenEye& 9801 (1 \%)\\  
         DOS attacks-Hulk& 107373 (10.8 \%)\\  
         DOS attacks-SlowHTTPTest& 32673 (3.3 \%)\\  
         DOS attacks-Slowloris& 2642 (0.3 \%)\\  
         FTP-BruteForce& 44816 (4.5 \%)\\  
         Infiltration& 21631 (2.1 \%)\\  
         SSH-Bruteforce& 43512 (4.3 \%) \\\bottomrule
    \end{tabular}
    \caption{Summary of the various attacks in the dataset.}
    \label{tab:data}
\end{table}
We use the Cyber Defense Dataset (CSE-CIC-IDS2018)~\cite{Dataset2018,Iman2018}, an Intrusion Detection System (IDS) dataset, from a collaborative project between the Communications Security Establishment (CSE) and the Canadian Institute for Cybersecurity (CIC). It contains 9 classes of network attacks and a total of 1,048,576 data points. The data was collected from an infrastructure that includes 420 machines and 30 servers. The dataset also includes network traffic and system logs. It uses CICFlowMeter to extract the features from the traffic logs\cite{Arash2017CICFlowMeter, Gerard2016CICFlowMeter} and contains a total of 80 statistical features, including flow duration, number of packets, length of packets, etc. Table~\ref{tab:data} summarizes the attacks and the number of data points in each class.

\subsection{Experimental Setup}
To train our model, we split our dataset into training (80\%) and test dataset (20\%).
For simplicity, we evaluate our approach with two cohorts with distinct privacy requirements and non-iid data distribution. To do so, we first randomly divide the dataset into two disjoint label sets, and each cohort is assigned a label set. This ensures that there is no overlapping of labels between the cohorts. We note that the label assignment to cohorts is performed multiple times, and we report the performance over multiple runs. 
Next, we assign the data samples to each client within a cohort. In this case, we split the number of samples uniformly across the clients, such that each client is assigned two labels. We note that the benign label is assigned to all clients. Thus, each client has no more than three labels assigned.
That is, the client has a benign label and two intrusion label.

We implemented our code using the Tensorflow framework~\cite{abadi2016tensorflow}. Our intrusion detection network consists of three layers; the input layer has 79 nodes, followed by two hidden layers of 79 and 128 nodes. The output layer has 9 nodes with softmax activation. We use sparse categorical cross-entropy loss to compute the training loss and Adagrad optimizer with a learning rate of 0.1. Moreover, we use a batch size of 10 in our experiments. 

Unless stated otherwise, and similar to~\cite{geyer2017differentially}, we use $\epsilon=6$ and $\epsilon=8$ for the two cohorts and $\delta=10^{-5}$. Moreover, we set the sensitivity $S$ parameter and the GM parameter $\sigma$ to 1. We run our experiments for 10,000 clients, with the random sub-sampling ratio set to 5\%, although we also perform sensitivity analysis with other values. In addition, we use the moments account to track the privacy loss, and the training stops when the privacy budget for both cohorts is spent. 
We use $\gamma=1$  and $\rho=0.25$ for our continual learning algorithms DP-R and DP-SI,  respectively.

\subsection{Baseline Techniques}

We compare our approach with two baseline techniques --- namely federated (non-private) and DP-federated algorithm. In the non-private scenario, we use the federated algorithm without privacy requirements. The data is distributed based on the process discussed above. However, the privacy parameter $\epsilon$ at each cohort is ignored. We train the federated model until the model converges.  
Note that the non-private technique provides an upper bound on the performance that a federated model can achieve.    
We also compare our approach with DP-Federated technique~\cite{geyer2017differentially}. For training the model, we use the same process discussed above. That is, the training with a cohort's gradient stops when their privacy budget is over.

\subsection{Evaluation Metrics}
To evaluate the performance, we use F1-score as our metric since we have imbalanced classes. In particular, we report the micro, macro and weighted-F1 score. We note that the micro F1-score measure the aggregate contribution of all classes. Thus, it treats each class label equally and does not give advantage to rarer classes. 
The macro F1-score computes the metric independently for each class and finds the average, thus, ignoring the label imbalance. 
On the other hand, the weighted F1-score calculates the metric for each label but finds their weighted average, accounting for the label imbalance. 
\begin{eqnarray*}
  \text{micro-F1} =& \text{F1-score}_{class 1+2 +...+N} \\
  \text{macro-F1} =& \frac{1}{N}\sum_{i}^N \text{F1-score}_i \\
    \text{weighted-F1} =& \sum_{i}^N w_i \cdot \text{F1-score}_i 
\end{eqnarray*}
where $w_i$ is the weight of the $i^{th}$ class that depends on the number of samples such that $\sum_{i} w_i = 1$.

%% file: evaluation.tex
In this section, we extensively evaluate our approach and compare it to other baseline approaches. 
We also analyze the sensitivity of our approach to various hyperparameters.  

\subsection{Performance comparison}

\begin{figure}
        \centering
        \begin{subfigure}[b]{0.24\textwidth}
            \centering
              \includegraphics[width=\textwidth]{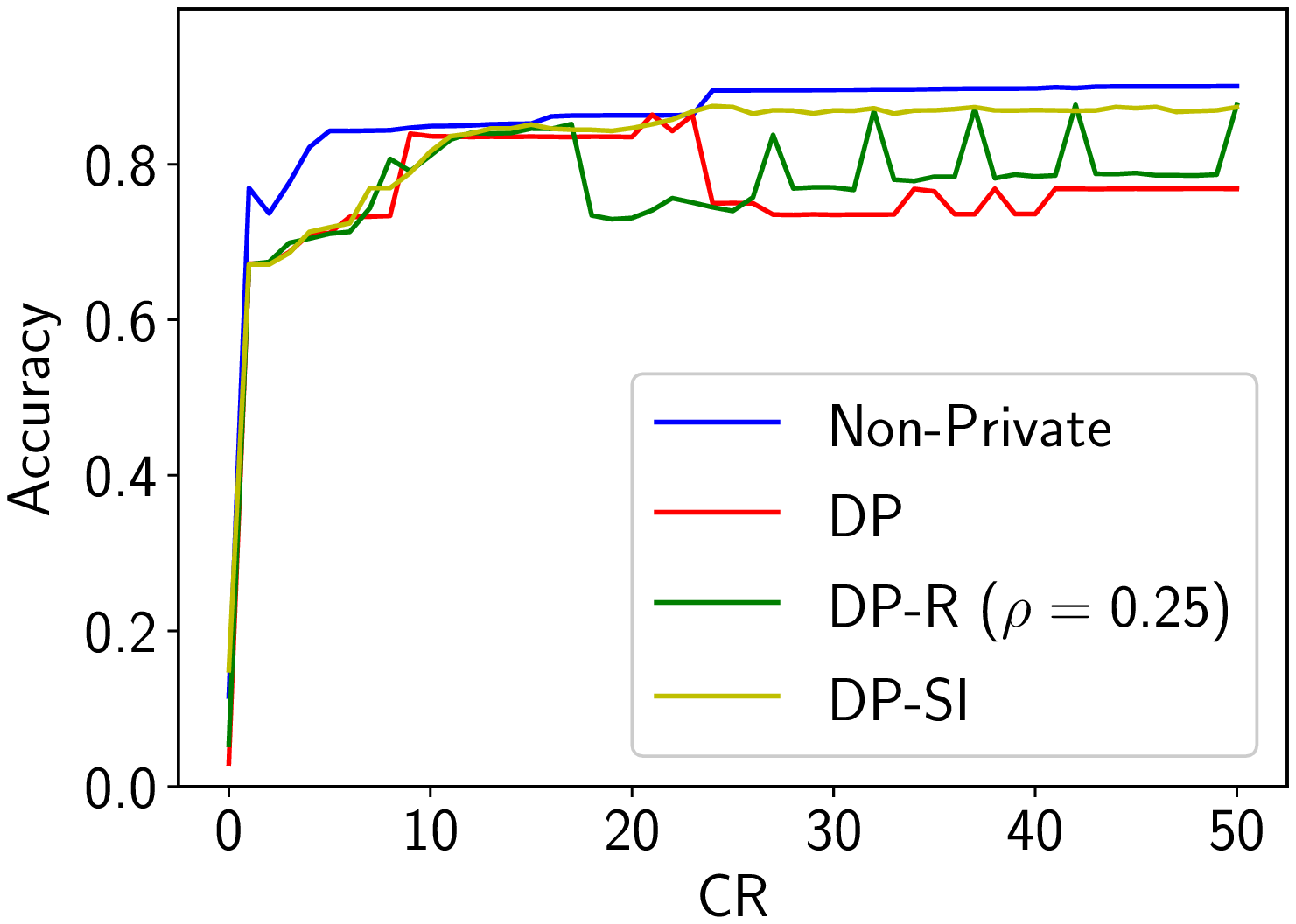}
            \caption[]%
            {{\small Training Accuracy}}    
            \label{fig:CLvsBase}
        \end{subfigure}
        \begin{subfigure}[b]{0.24\textwidth}  
            \centering 
            \includegraphics[width=\textwidth]{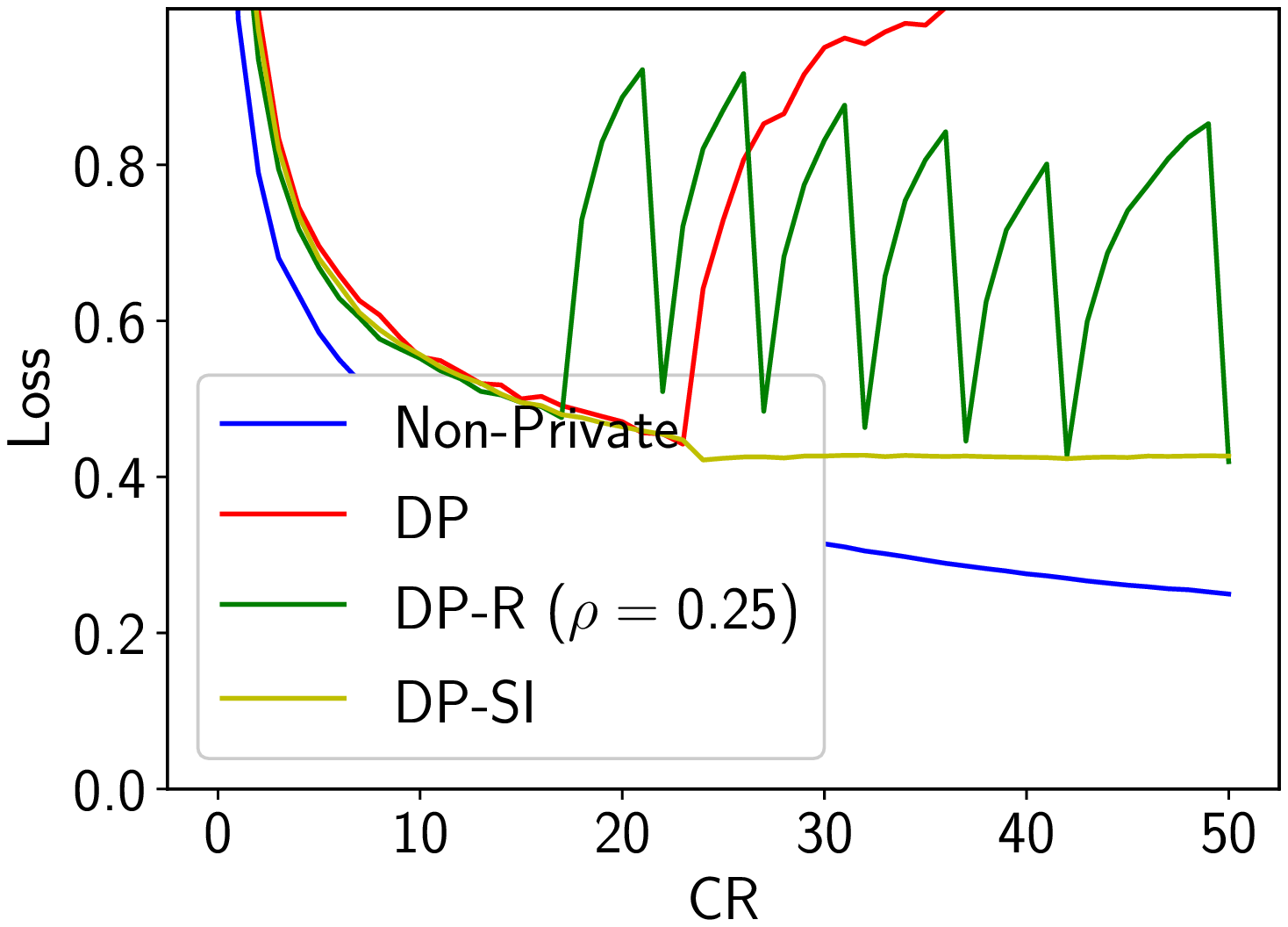}
            \caption[]%
            {{\small Training Loss}}    
            \label{fig:CLvsBaseLoss}
        \end{subfigure}
        \caption[]
        {\small Training performance for varying communication rounds.}
        \label{fig:training}
\end{figure}

We first evaluate the training performance of our approach with other baseline techniques. 
To do so, we divide the devices into two cohorts, cohort 1 and 2, with $\epsilon=6$ and $\epsilon=8$, respectively and $\delta=10^{-5}$. We then train the model until the privacy budgets of the cohorts are spent. We use this setup to train the differentially private models.
In the non-private scenario, we do not assume any privacy budget and train using the federated approach until the model converges.  

Figure~\ref{fig:training} depicts the performance of different approaches during the training period. We observe that the federated (non-private) scenario achieves the maximum accuracy (and the lowest loss value) and provides an upper bound on the maximum performance the intrusion detection model can achieve. On the other hand, the differentially private approach (without continual learning) achieves the lowest overall performance among all the techniques at the end of the training period. This is because the DP-only approach \emph{forgets} past experiences when the privacy budget of cohort 1 is spent. As shown, at CR={23}, we see a significant drop in accuracy (and loss) in the DP-only approach, where the privacy budget of cohort 1 finishes and all new updates to the model come from cohort 2, which results in the drop. In comparison, the performance of our continual learning-based DP approach remains stable during the training phase even when the privacy budget is spent. 

Note that in DP-Rehearsal (DP-R) $\rho=0.25$ indicates 25\% of cohort 1's gradient updates are uniformly spread throughout the training phase, resulting in a sawtooth-like pattern. This is because the updates from cohort 1 at regular intervals prevents the model parameters from deviating. We observe that the performance of DP-Synaptic intelligence (DP-SI) is more stable as it minimizes significant changes to the model parameters from subsequent updates, which allows it to retain experiences learned from previous tasks. 

\begin{table}[t]
\centering
\caption{Performance comparison with other baseline techniques.}
\label{tab:Performance_table}
\begin{tabular}{l|ccc}
    \toprule
    Model  & Macro-F1 &  Weighted-F1 & Micro-F1\\ \midrule
	Non-Private&0.77&0.94&0.95\\
	DP&0.38&0.74&0.81\\
	DP-R ($\rho = 0.25$)& {\bf 0.49}&{\bf 0.85}&{\bf 0.88}\\
	DP-SI&0.46&0.85&0.87\\
    \bottomrule
\end{tabular}.
\end{table}

Next, we compare the efficacy of our approach with other baseline techniques.  
Table~\ref{tab:Performance_table} summarizes the performance of different approaches on the test dataset in classifying various types of anomalies. Note that the federated (non-private) approach provides an upper bound on the performance. We observe that the differentially private approach achieves an F1 score of 0.81. We note that our continual learning-based approach outperforms the differentially private (DP) approach. In addition, both DP-Rehearsal and DP-SI achieves similar performance, with the micro-F1 score equal to 0.88 and 0.87, respectively and weighted-F1 score being the same. 

In summary, our proposed methods DP-SI and DP-R outperform the baseline DP method in all metrics. It is important to note, that the macro-F1 of all models is pushed lower due to the inherent class imbalance in our data. However, our proposed methods outperform the baseline federated DP method in macro averaged scores as well.

\subsection{Model Flexibility}

\begin{figure}
        \centering
        \begin{subfigure}[b]{0.24\textwidth}
            \centering
        \includegraphics[width=\linewidth]{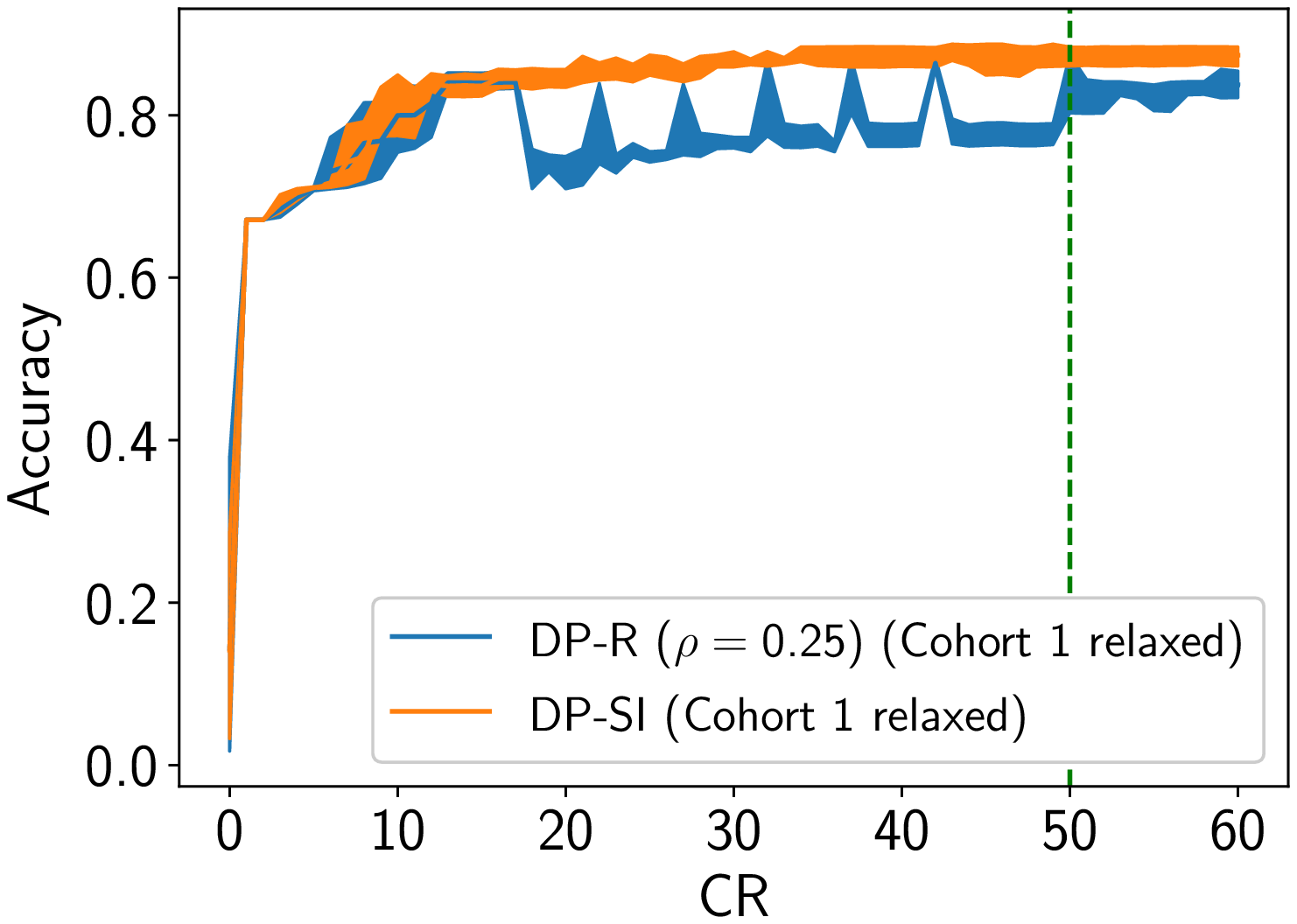}
            \caption[]%
            {{\small Cohort 1}}    
            \label{fig:Rehearsal_SI_CR1}
        \end{subfigure}
        \begin{subfigure}[b]{0.24\textwidth}  
            \centering 
        \includegraphics[width=\linewidth]{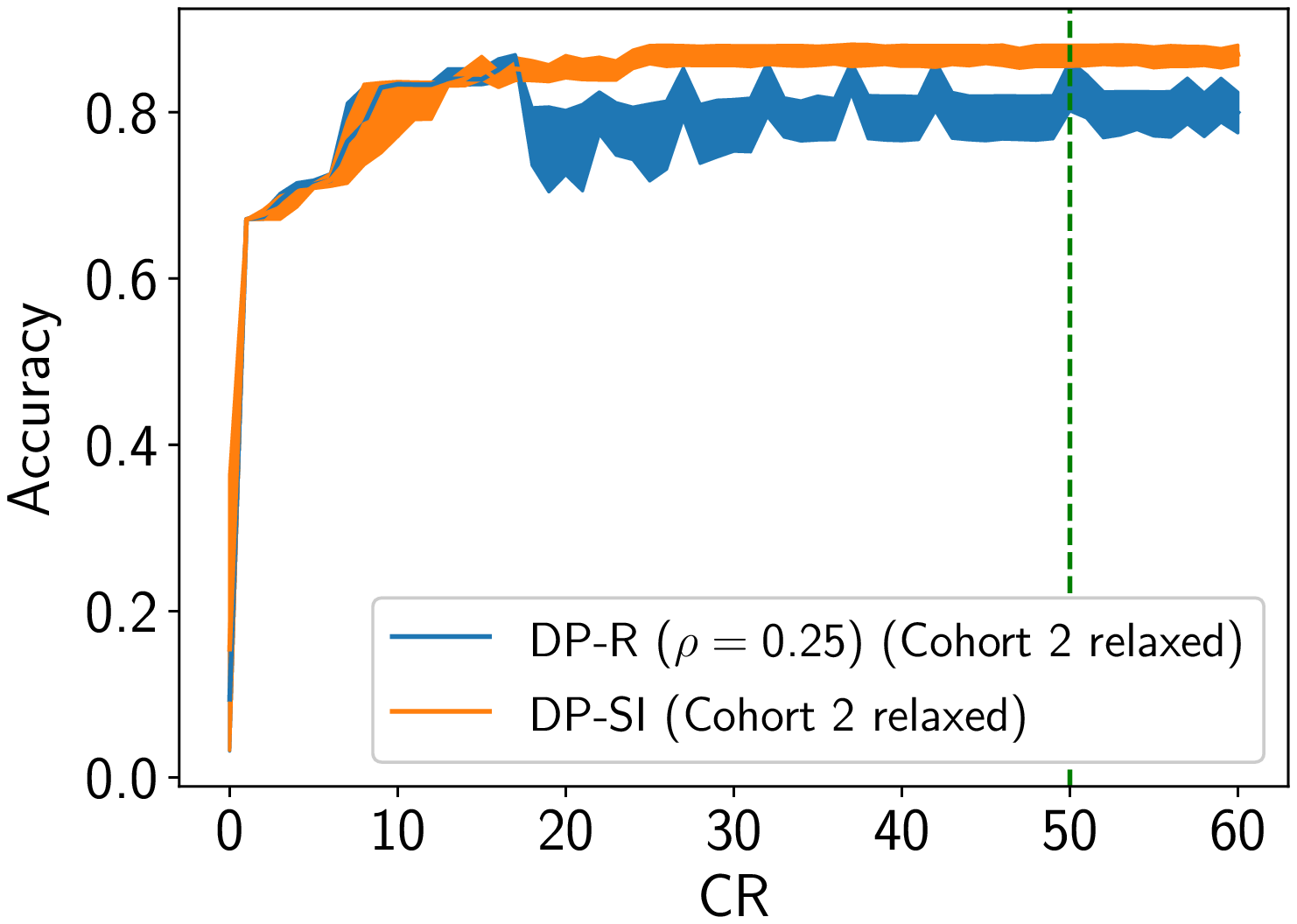}
            \caption[]%
            {{\small Cohort 2}}    
            \label{fig:Rehearsal_SI_CR2}
        \end{subfigure}
        \caption[]
        {\small Training flexibility comparison between DP-Rehearsal and DP-SI. The privacy budget of a cohort is relaxed to enable further training.}
        \label{fig:flex}
\end{figure}

Next, we compare the training flexibility of DP-Rehearsal and DP-SI. For this experiment, we consider the scenario where the trusted curator decides to \emph{relax} the privacy requirements of a cohort after the privacy budgets of both cohorts are over. That is, the trusted curator continues to train the model with new updates. In the real-world, this may be performed to allow new updates to the model. We do so by allowing the framework to update the model with a cohort's gradient for an additional ten communication round to indicate a relaxation of privacy of the cohort.

\begin{table}[t]
\centering
\caption{Performance comparison after relaxing privacy in case of continual learning}
\label{tab:Privacy_relaxed}
\begin{tabular}{l|ccc}
    \toprule
    Model  & Macro-F1 &  Weighted-F1 & Micro-F1\\ \midrule
    DP-SI Cohort 1 Relaxed &{\bf 0.49}& {\bf 0.85}& {\bf 0.87}\\
	DP-R Cohort 1 Relaxed&0.42&0.78&0.83\\
	DP-SI Cohort 2 relaxed&{\bf 0.46}& {\bf 0.85}&{\bf 0.87}\\
	DP-R Cohort 2 Relaxed&0.39&0.71&0.79\\
    \bottomrule
\end{tabular}.
\end{table}

Figure~\ref{fig:flex} depicts the training performance when we relax the privacy budget of cohort 1 and cohort 2 independently.
As shown in the figure, at CR=50, the overall training accuracy of DP-Rehearsal drops when we update the model with new information. When the privacy budgets are spent in DP-Rehearsal, only updates from one cohort are available, resulting in the model forgetting previously learned tasks. However, in DP-SI, the training performance remains stable and doesn't drop even when the privacy budget is spent. This indicates that DP-SI is more flexible and allows the privacy budget to change in the future without adversely impacting model performance. As the figure shows, the model performs consistently better regardless of which cohort's budget is relaxed. Table~\ref{tab:Privacy_relaxed} summarizes the overall performance on the test dataset and shows it outperforms DP-Rehearsal when the privacy budget of a cohort is relaxed during training.

\begin{figure}
    \centering
    \includegraphics[width=2.5in]{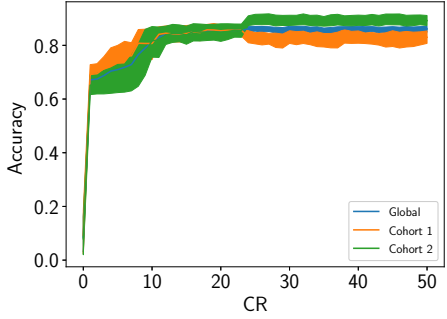}
    \caption{ Cohort-wise training accuracy of DP-SI ($\gamma=1$). } 
    \label{fig:groupwise}
\end{figure}

We also evaluate the DP-SI model's performance during training on cohort 1 and cohort 2 individually and compare it to the overall accuracy. 
Note that the data among the cohorts is non-iid, i.e., the cohorts' data distribution is distinct. 
As again, we note that the model's accuracy on cohort 1 (with lower epsilon value) decreases during the training period when the privacy budget of cohort 1 ends at CR=23. But, because synaptic intelligence constrains the loss function, it reduces the effects of applying the new gradient updates. We also observe an increase in overall training accuracy of cohort 2 due to increased update. However, as shown, the model converges even with additional updates from cohort 2. 

From these results we can conclude that DP-SI provides more flexibility to the end-user in terms of changes to privacy considerations than DP-R.

\subsection{Impact of Hyperparameters}

\begin{figure}
        \centering
        \begin{subfigure}[b]{0.24\textwidth}
            \centering
        \includegraphics[width=\textwidth]{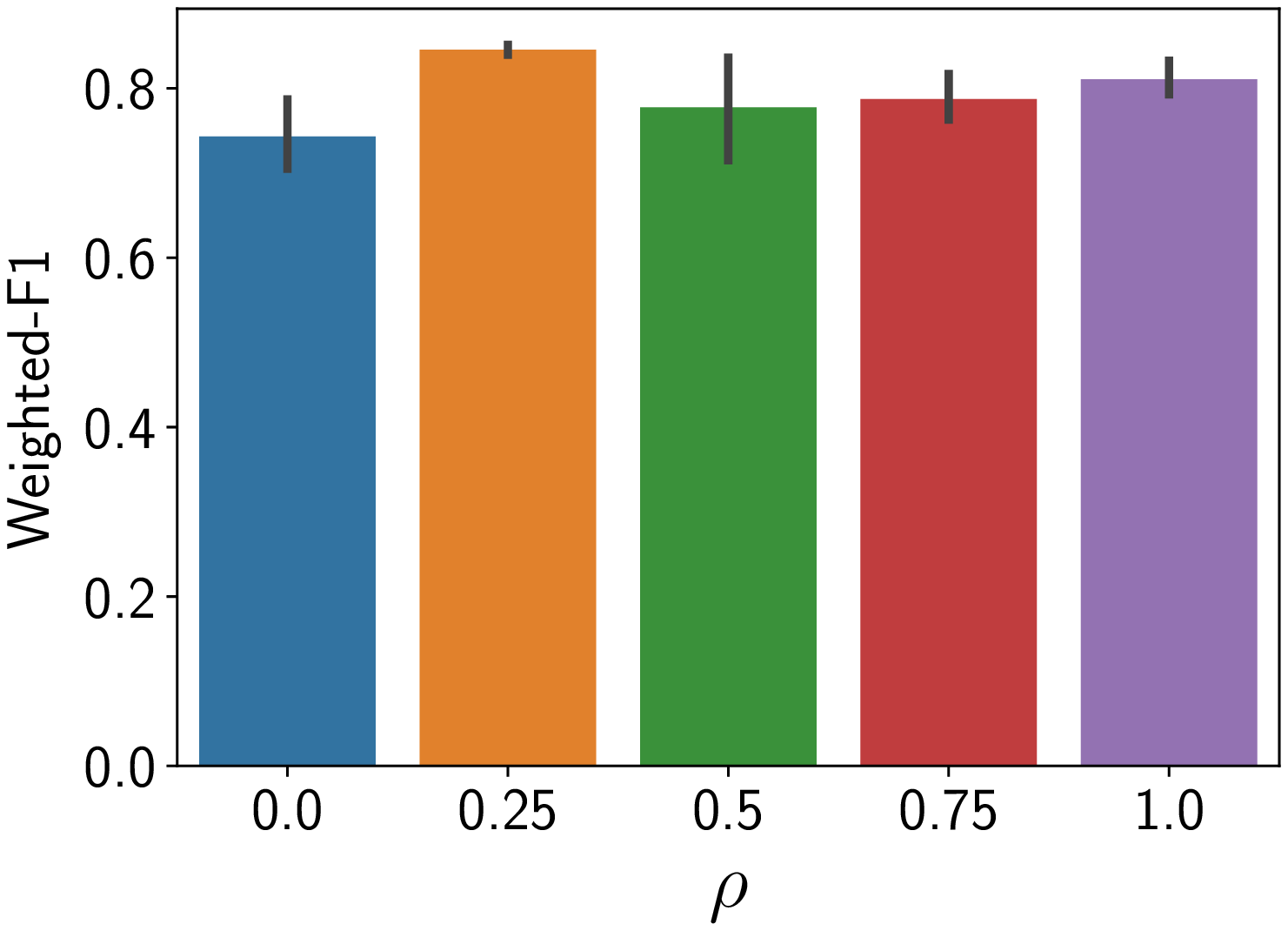}
            \caption[]%
            {{\small DP-Rehearsal}}    
            \label{fig:rehearsal}
        \end{subfigure}
        \begin{subfigure}[b]{0.24\textwidth}  
            \centering 
        \includegraphics[width=\textwidth]{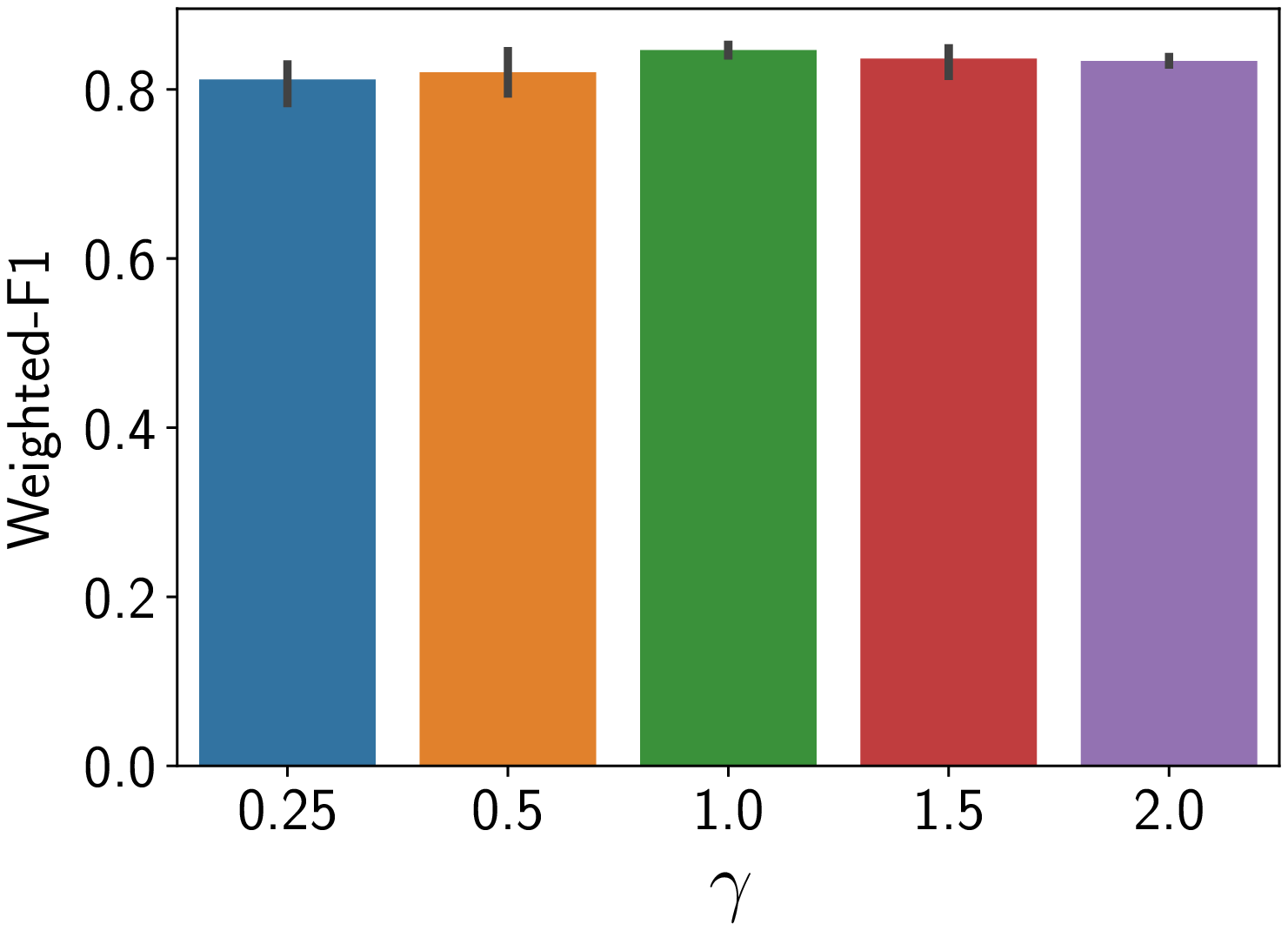}
            \caption[]%
            {{\small DP-SI}}    
            \label{fig:gamma}
        \end{subfigure}
        \caption[]
        {\small Effect on performance for varying hyperparameters}
        \label{fig:hyperparam}
\end{figure}

We evaluate the impact of DP-Rehearsal parameter ($\rho$) that controls how the gradient updates are distributed during the training phase. Recall that ($\rho=0.5$) indicates that half the privacy budget is used in the first half of the training phase, and the next half of the privacy budget is uniformly distributed for the rest of the training phase to ensure that the model is periodically updated. Similarly, when ($\rho=0$), all the privacy budget is used up at the start, and thus, this scenario defaults to the DP-only scenario.  

Figure~\ref{fig:rehearsal} shows the performance of DP-Rehearsal for various $\rho$ values across multiple runs. As shown, when $\rho=0$, it has the lowest F1 score, as the gradient updates of cohort 1 is not used for future model updates. Moreover, in our evaluation, we find that DP-Rehearsal performs best when $\rho=0.25$. In practice, however, we believe an extensive grid search is required to determine the best model.  

Figure~\ref{fig:gamma} shows the performance of DP-SI for varying $\gamma$ values. Recall that $\gamma$ controls the tradeoff between the past and future experiences learned by the model. Thus, when $\gamma<1$, the model provides more weight to newer experiences, whereas $\gamma>1$ gives more weight to past experiences. And, $\gamma=1$ corresponds to equal weight to past and new experiences. We note that the model performs best when $\gamma=1$. However, when we increase the preference to either past or future experience by increasing or decreasing $\gamma$, respectively, we observe a slight decrease in performance. In particular, we observe that when $\gamma=1$, the DP-SI's F1-score is 0.846, and 0.811 and 0.812 for $\gamma=0.25$ and $\gamma=2.0$, respectively.  

While our methods are sensitive to their relevant hyperparameters and a grid search is recommended before deployment, it should be noted that all hyperparameter configurations of DP-SI outperform the baseline DP model. Additionally, DP-R hyperparameter combinations except $\rho = 0$ also outperform the baseline (DP-R with $\rho = 0$ is identical to the baseline).

\subsection{Impact of Sub-sampling Clients}

\begin{figure}
        \centering
        \begin{subfigure}[b]{0.24\textwidth}
            \centering
        \includegraphics[width=\textwidth]{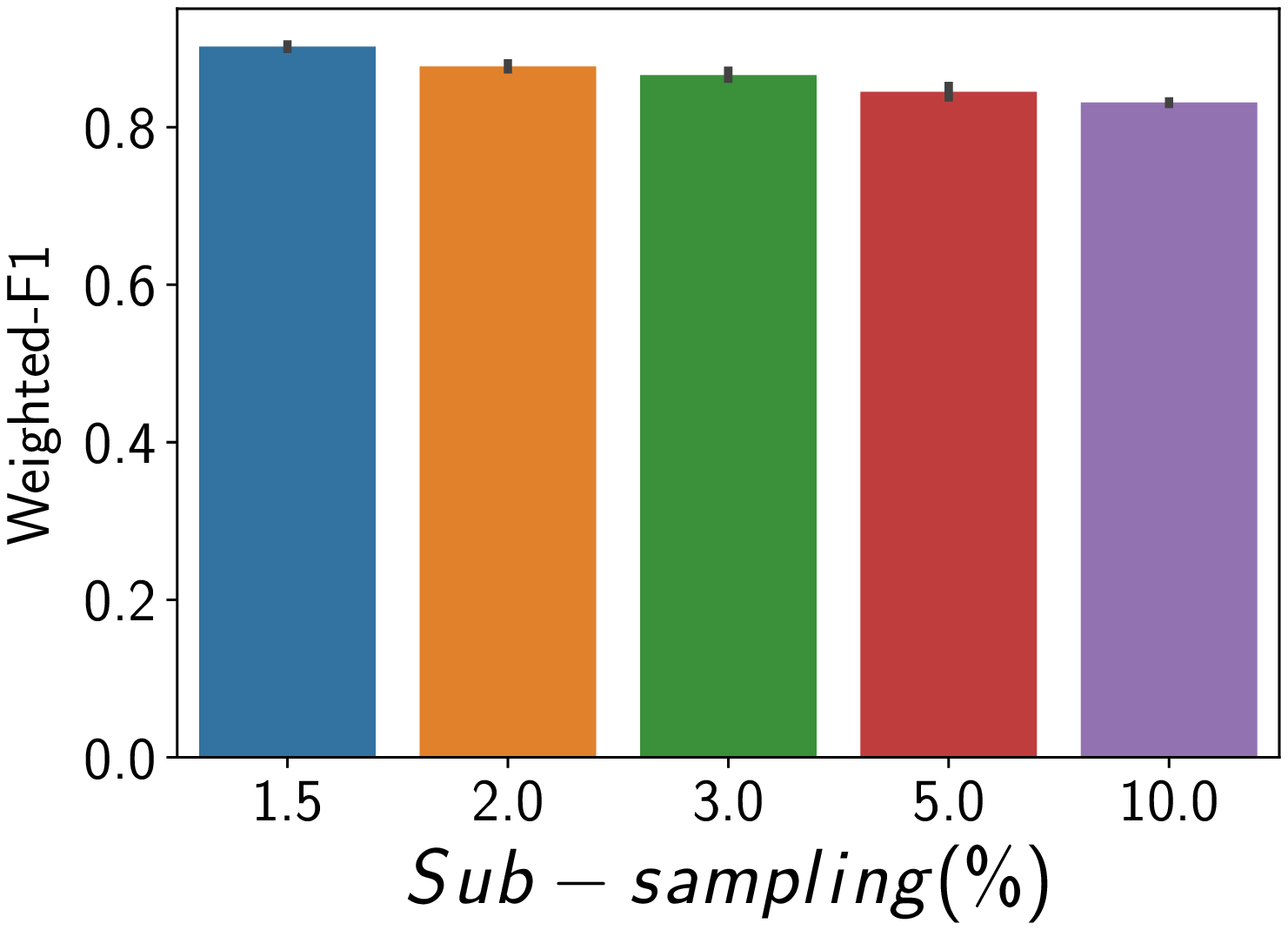}
            \caption[]%
            {{\small F1-score}}    
            \label{fig:subsampling}
        \end{subfigure}
        \begin{subfigure}[b]{0.24\textwidth}  
            \centering 
        \includegraphics[width=\textwidth]{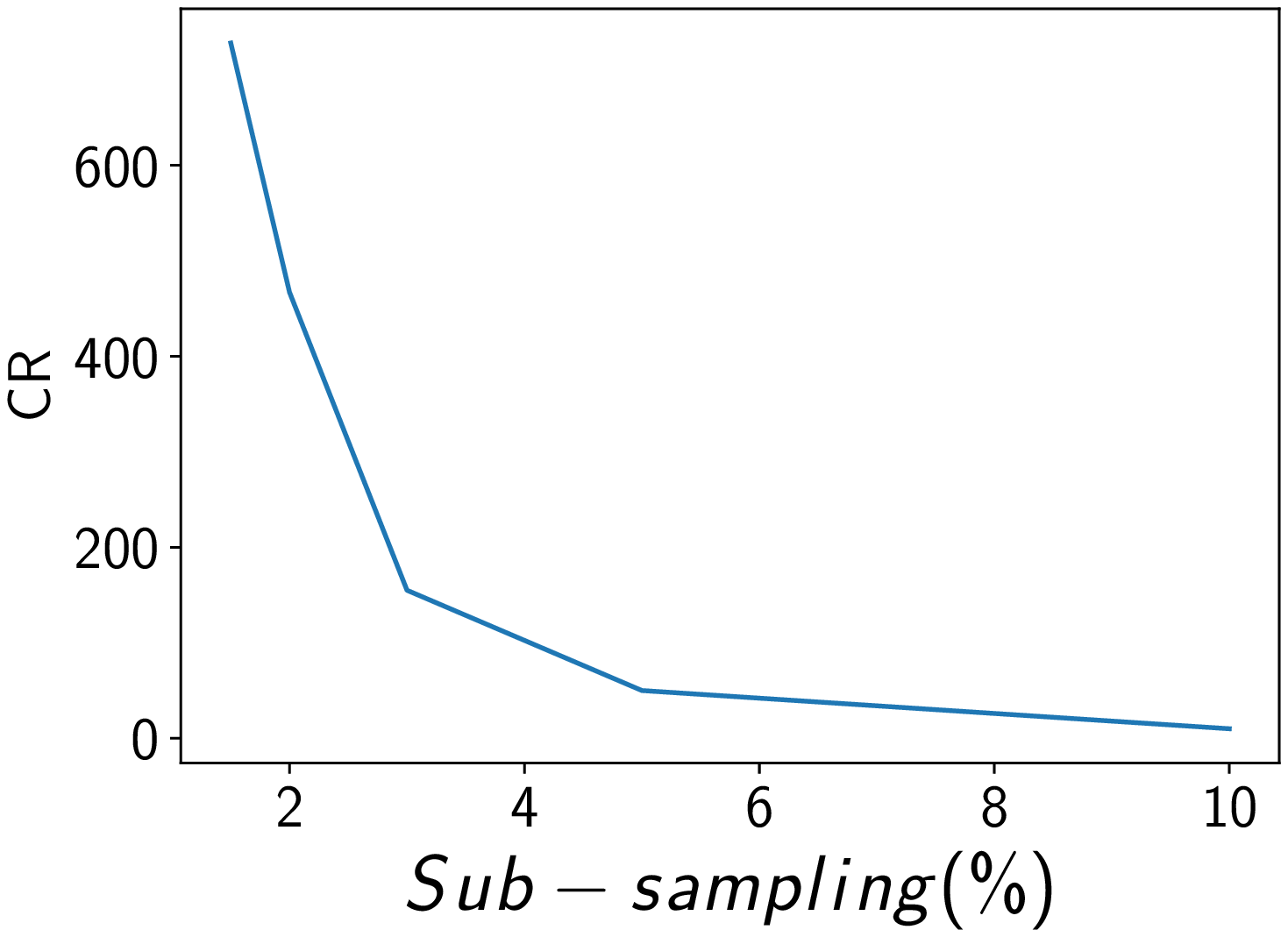}
            \caption[]%
            {{\small Communication rounds}}    
            \label{fig:cr}
        \end{subfigure}
        \caption[]
        {\small Effect of different sub-sampling values.}
        \label{fig:privacyamp}
\end{figure}

We now evaluate the effect of varying sub-sampling values on the overall performance.  Note that the sub-sampling ratio amplifies privacy~\cite{balle2018privacy}. That is,  differentially private mechanism provides much better privacy guarantees when performed on a random subsample of the clients.  
So far, we have used randomly sampled 5\% of the clients at each communication round.
Figure~\ref{fig:subsampling} shows the performance of DP-SI for varying sub-sampling percentages. We observe that as the sub-sampling percentage increases, the accuracy of the detection model decreases. In other words, the model performance improves when a small fraction of the clients participate in each communication round. In particular, we observe that when sub-sampling  decreases from 1.5\% to 10\%, the F1-score decreases by 7.89\%. This can also be seen in Figure~\ref{fig:cr}, where the number of communication round increases with smaller sub-sampling values. As shown, the communication round decreases from  728 to 10, when sub-sampling value changes from 1.5\% to 10\%.

%% file: future.tex
In this work, we simulate a multi-cohort system and use it to evaluate the performance of federated DP-SGD methods on client cohorts with heterogeneous privacy budgets and nonidentical data distributions. While our experiments have focused on a 2-cohort proof-of-concept setting, our gradient reuse mechanism for DP-SI and periodic rehearsal mechanism for DP-R are defined for any number of client-cohorts. Therefore, our methods can be used for any $N$ client-cohort system. Additionally, CL methods like Rehearsal and SI have been shown \cite{li2017learning,zenke2017continual}  to improve model performance in experiments with more than two tasks. Therefore, we expect the advantages provided by our methods to generalize well to $N$ client-cohort systems. 

A possible shortcoming of our model is tied to the inherent limitations of an $(\epsilon,\delta)$-DP compliant system. DP algorithms reduce and limit the affect of any single data sample on the model~\cite{dwork2014algorithmic}. As a result, $(\epsilon,\delta)$-DP trained systems may have disproportionately reduced performance on rarer classes that have a very small number of training samples. We observe similar behavior in our experiments wherein the DP limits the influence from rarer classes. DP-SGD baseline, DP-SI and DP-R models have trivial performance on the rarest classes such as DOS attacks-Slowloris (see Table \ref{tab:data}). Investigating methods that can leverage  our problem structure to improve rare class performance remains part of our future work.

%% file: relatedwork.tex
Existing research related to our work can be categorized as follows.
\\
{\bf Learning and Differential privacy: }
Differential privacy was proposed by Dwork et al.\cite{dwork2014algorithmic}, and several works followed that combined machine learning with Differential privacy (DP)~\cite{abadi2016deep, wang2019subsampled, mironov2017renyi}. These approaches account for the privacy loss during training the model. For example, in \cite{abadi2016deep}, authors use moments accountant to estimate the privacy loss using higher moments of the privacy loss random variable. Our work uses the moments accountant for tracking privacy loss.

There has also been much work on federated learning~\cite{mcmahan2017communication, Kairouz2019, zhao2018federated, huang2020dp, nguyen2019diot}. A federated learning approach is useful, especially when data is distributed, and explicit data sharing is undesirable for privacy or regulatory reasons. In~\cite{mcmahan2017communication} proposed Federated Averaging algorithm to learn a shared model in a decentralized manner.  Separately, \cite{huang2020dp} proposed DP-federated learning for unbalanced data, where the amount of data available at each client  differs.
Recently, \cite{geyer2017differentially} proposed DP-federated learning at the client-level, which provides privacy guarantees at the client-level instead of data samples level. Our approach adapts this client-level framework to provide differential privacy at a cohort-level. However, unlike prior work, we consider heterogeneous privacy budgets among the cohorts.

The concept of users with heterogeneous privacy has been previously proposed in~\cite{alaggan2015heterogeneous, jorgensen2015conservative}. 
In \cite{jorgensen2015conservative}, the authors define the notion of personalized differential privacy by providing user or record specific privacy requirements. Similarly, heterogeneous differential privacy was formally defined in~\cite{alaggan2015heterogeneous} to provide user-item specific privacy requirements. 
There have also been other efforts to model user-specific privacy requirements\cite{li2017partitioning,doudalis2017one,tian2017novel, palanisamy2017group}.
\cite{li2017partitioning,doudalis2017one,tian2017novel}. For instance, \cite{li2017partitioning} provides partitioning mechanism to group users with personalized privacy requirements into different $\epsilon$ partitions under a non-federated setting. 
Separately, there has been much work on continual learning, where the key idea is to accumulate knowledge without forgetting previously learned behavior~\cite{kirkpatrick2017overcoming,lopez2017gradient,chaudhry2018efficient,zenke2017continual,li2017learning}. 
In the same way, we use the continual learning technique to ensure that the model remembers past experiences when the privacy budget of a cohort ends.  
To the best of our knowledge, it is the first system that studies continual learning within the DP-federated framework.

{\bf Intrusion Detection:}
There has been significant work on using network traffic to detect intrusions~\cite{sekar2002specification,raza2013svelte,sommer2010outside}. With advances in machine learning, recent approaches have proposed deep learning-based approaches for network anomaly detection~\cite{mirsky2018kitsune, jia2017contexlot}. For example, in~\cite{mirsky2018kitsune}, authors proposed autoencoders to differentiate normal from anomalous traffic. 
Separately, there has been work that studies anomaly detection in a private manner~\cite{asif2019accurately}. 
However, prior work mostly assumes homogeneous privacy budgets. These approaches do not work in the context of heterogeneous privacy budgets. Moreover, we propose a differentially private CL approach that mitigates the issues from heterogeneity.

%% file: conclusion.tex
In this paper, we investigated detecting network intrusions in IoT devices for users with heterogeneous privacy requirements and nonidentical data distribution. 
We argued users may be grouped into cohorts and presented the notion of cohort-based federated differential privacy. The cohort-based differential privacy approach tracks the privacy requirements for each cohort.  Moreover, we showed that traditional differential privacy approaches do not perform well when cohorts have different privacy requirements and nonidentical data distribution. To solve the problem, we proposed a novel continual learning-based federated DP approach that can handle such heterogeneous privacy budgets. We extensively evaluated our approach on a real-world intrusion dataset that consists of various attack vectors. Further, we demonstrated that our proposed approach outperforms traditional federated DP-based approaches. 
Our results also showed the efficacy of our approach in detecting intrusion patterns within the differential privacy framework.